%% file: iclr2026_conference_camera-ready.tex
\documentclass{article} 
\usepackage{iclr2026_conference,times}

\input{math_commands.tex}

\usepackage{hyperref}
\usepackage{url}
\usepackage{graphicx} 
\usepackage{wrapfig}
\usepackage{booktabs}
\usepackage{multirow}
\title{Toward World Models for Epidemiology}

\author{
\hspace{-0.5em}
\textbf{Zeeshan Memon}$^{1,}$\thanks{Equal contribution.},
\textbf{Yiqi Su}$^{2,}$\footnotemark[1],
\textbf{Christo Kurisummoottil Thomas}$^{3}$,
\textbf{Walid Saad}$^{4}$, \\
\textbf{Liang Zhao}$^{1}$,
\textbf{Naren Ramakrishnan}$^{2}$ \\
$^{1}$Department of Computer Science, Emory University, Atlanta, GA, USA \\
$^{2}$Department of Computer Science, Virginia Tech, Alexandria, VA, USA \\
$^{3}$Department of Electrical and Computer Engineering, Worcester Polytechnic Institute, Worcester, MA, USA \\
$^{4}$Bradley Department of Electrical and Computer Engineering, Virginia Tech, Alexandria, VA, USA \\
\texttt{\{zeeshan.memon, liang.zhao\}@emory.edu} \\
\texttt{\{yiqisu, walids, naren\}@vt.edu} \\
\texttt{cthomas2@wpi.edu}
}

\iclrfinalcopy 
\begin{document}

\maketitle

\begin{abstract}
World models have emerged as a unifying paradigm for learning latent dynamics, simulating counterfactual futures, and supporting planning under uncertainty. In this paper, we argue that computational epidemiology is a natural and underdeveloped setting for world models. This is because epidemic decision-making requires reasoning about latent disease burden, imperfect and policy-dependent surveillance signals, and intervention effects are mediated by adaptive human behavior. We introduce a conceptual framework for epidemiological world models, formulating epidemics as controlled, partially observed dynamical systems in which (i) the true epidemic state is latent, (ii) observations are noisy and endogenous to policy, and (iii) interventions act as sequential actions whose effects propagate through behavioral and social feedback.
We present three case studies that illustrate why explicit world modeling is necessary for policy-relevant reasoning: strategic misreporting in behavioral surveillance, systematic delays in time-lagged signals such as hospitalizations and deaths, and counterfactual intervention analysis where identical histories diverge under alternative action sequences.
\end{abstract}

\section{Introduction}

World models are growing in popularity across multiple domains in machine learning. Bridging latent dynamics modeling, imagination-based planning, and counterfactual rollouts, the core idea is to learn an internal representation of how the world evolves and then use that representation to support decision-making~\cite{wm}. Such learned simulators have been deployed for control, robotics, game playing, and embodied agents, including Dreamer-style agents and MuZero-like systems~\cite{hafner2020, schrittwieser2020mastering}. At their core, world models provide the glue among perception, action, and reasoning: they separate latent state from noisy observations, thereby enabling prediction, simulation, and counterfactual reasoning at scale~\cite{silver2021reward}.

A typical world model includes several ingredients. First, it is assumed to have a latent state representation that compresses the relevant aspects of the environment into a compact internal variable~\cite{kaelbling1998planning}. Given the state representation, we also need transition dynamics that describe how this latent state changes over time, often as a function of actions or interventions~\cite{hzl2025identifying, wang2024representation}. World models also include an observation function that maps latent states back to the measurements we actually see, capturing noise, reporting artifacts, and partial observability~\cite{lillemark2026, li2025survey}. Expanding further, we aim to support a notion of interventions or actions, i.e., what can be changed, when, and how those changes propagate through the system. Finally, a world model is trained and evaluated against objectives or criteria that reflect the intended use, such as predictive accuracy, planning performance, or the ability to generate plausible counterfactual rollouts under alternative decisions~\cite{wm,ke2021systematic}.

\paragraph{Why world models for epidemiology?}
In this paper, we investigate world models for computational epidemiology. Computational epidemiology has undergone a major transformation over the last two decades (even prior to COVID-19)~\cite{salathe2012digital, o2017digital}. Early work centered on compartmental, mass-action models with SEIR-style dynamics: analytically tractable, interpretable, and naturally aligned with policy questions~\cite{Kong2022}. Over time, the data landscape changed through new surveillance modalities, including search trends, digital traces, mobility data, social media, and other forms of syndromic surveillance, which in turn encouraged machine learning approaches layered on top of mechanistic epidemiological modeling.

The biggest beneficiary of this shift has been forecasting~\cite{tabataba2017framework}. Short-term case prediction, hospital demand estimation, and ensemble forecasting have become central, with forecast accuracy often treated as the dominant evaluation metric. But the field has increasingly moved beyond forecasting toward scenario analysis~\cite{chen2020scenario}, wherein the goal is to support questions from policy makers such as: should we close schools? should we quarantine a sub-population? and what to expect under alternative interventions? This naturally invites counterfactual rollouts of policies and stress-testing of non-pharmaceutical interventions, vaccination strategies, and reopening plans~\cite{Davies2020NHB,Ferguson2020Imperial}.

However, much of current scenario analysis research still assumes that the world is given, i.e., fixed dynamics, fixed behavior, and fixed observation processes~\cite{COUZIN2023}. Data are often treated as exogenous, and humans are treated as parameters rather than participants in the system being studied~\cite{KRAKOWSKI2025}. From a world model perspective, these assumptions are limiting, because they conflate the underlying latent epidemic state with shifting observation mechanisms and adaptive human behavior, and they restrict counterfactual reasoning to interventions applied to an otherwise static simulator~\cite{wm, Fenichel2011}.

Our contributions are three-fold: (i) a formal
conceptual framework for epidemiological world
models, (ii)
reframing of epidemic decision making
into reasoning with world models,
which in turn supports
robustness of decisions under partial observability, endogenous measurement, and model misspecification, (iii)
three empirical case studies from our research that demonstrate the kinds of policy reasoning world models make possible.

\section{Some Epidemiology for Machine Learners}
This section serves as a `101’ overview that provides the epidemiological background needed for future sections.

\subsection{Compartmental Models}
\vspace{-2mm}
Compartmental models have historically been the backbone of epidemiological modeling~\cite{ChoiJTB2020}.
In this approach, the entire population under study is expected to be distributed across several states and dynamical rules described as a system of ODEs govern how people move across states. Classic examples include SIR (Susceptible-Infected-Recovered)~\cite{sir} and SEIR (Susceptible-Exposed-Infected-Recovered)~\cite{seir}. In their standard form, these models are expressed as systems of ordinary differential equations (ODEs) that describe how the fraction (or count) of the population in each compartment changes continuously over time~\cite{thiagarajan2022machine}. Such models can be enlarged with additional states that distinguish age structure, spatial structure, and risk strata~\cite{EARTH2025}. These models are often built around mass-action assumptions and 
rely on homogeneous mixing as a modeling convenience, even though real populations rarely mix homogeneously.

A primary virtue of compartmental models is their interpretability and policy relevance~\cite{Gostic2020}. Their parameters can often be treated as levers that correspond to actionable mechanisms, and summary quantities like $R_0$ (reproduction number) are commonly used to communicate epidemic status~\cite{r0, SMITH20123}. Informally, $R_0$ is the expected number of secondary infections caused by a single infected individual in a fully susceptible population, under a given set of behavioral and environmental conditions. 
When $R_0 = 1$, each infected person infects one other person on average, so incidence is expected to remain roughly steady rather than growing or shrinking. When $R_0 > 1$, infections grow because each case generates more than one new case on average, leading to expanding transmission. When $R_0 < 1$, infections decline because each case generates fewer than one new case on average, so transmission tends to die out over time.
In general, these reproduction numbers should be understood as summaries of system behavior rather than fixed constants~\cite{HEESTERBEEK20073}. Most importantly, interventions (by public health authorities) are imposed to bring $R_0$ to manageable levels.

\subsection{Agent-based models}
\vspace{-2mm}
Agent-based models provide a more fine-grained representation of epidemic spread by simulating individuals and their interactions~\cite{Tracy2018ABM, diffabm}. Here instead of ODEs, we model the population as a graph where nodes denote people and edges capture, e.g., contact structure. Once again nodes can be in various states (S/I/R or S/E/I/R, for instance). Time is typically discrete in these models and over the passage of time, people move from one state to another based on contact dynamics (e.g., if a susceptible node is exposed to a sufficient number of infected neighbors, the node moves to the exposed state in the case of SEIR). In an agent-based model we typically run simulations many times under the same model settings (often with different random seeds) to obtain a composite picture of disease spread across the network~\cite{rsif2024}.

\subsection{Observations as Noisy Proxies}
\vspace{-2mm}
In epidemiology, true incidence is typically latent. This is because infections may be asymptomatic, reporting may be delayed, and testing is often partial or uneven. As a result, observed signals are noisy proxies for the underlying epidemic state: reported cases are not the same as infections, hospitalizations tend to lag transmission dynamics, and deaths reflect historical state rather than current conditions~\cite{Dailey2007Timeliness,su2026epinode}. For a machine learning audience, the key distinction is between latent state and observed data streams~\cite{memon2025deep, Gostic2020}.

\subsection{Adversarial actors in epidemiology}
\vspace{-2mm}

Epidemiological systems can include adversarial or strategic behavior, especially with self-reported data from individuals. For example, a national study in Kenya found that while only 12\% of individuals reported not wearing masks, direct observation revealed that nearly 90\% were non-compliant, a discrepancy exceeding 75 percentage points that was consistent across demographic groups \cite{kenya2021}. Similar survey-based studies in countries like the USA document substantial social-desirability and recall biases in self-reported masking and vaccination behavior, compounded by spatial heterogeneity, policy variation, and limited auditability of daily behaviors \cite{spatiotemporal}.

\subsection{Data regimes and surveillance evolution}
\vspace{-2mm}
Traditional surveillance systems have relied on lab-confirmed cases and sentinel networks. More recently, surveillance has expanded to include digital and opportunistic data streams such as mobility traces, search activity, social media signals, and wastewater measurements~\cite{ZHANG2024, salathe2018digital}. In these modern regimes, data can be endogenous: policy affects behavior, behavior affects data, and data affects policy. As a result, the observation process itself can change under intervention, meaning that shifts in measured signals may reflect surveillance and behavioral feedback effects rather than true changes in underlying transmission.

\subsection{Limits of forecasting-centric thinking}
Forecasts are best understood as conditional extrapolations: they project observed trends forward under the assumptions encoded by the model and the data-generating process. As a result, forecast skill is not the same as policy usefulness. A model can produce accurate short-term predictions while still relying on wrong structural assumptions, and it can present misleading confidence when the observed signals are biased, delayed, or otherwise distorted~\cite{Yuan2021TemporalBias}. In particular, when we ask a high-performing model to adjust its forecasts under a change of scenario, 
(e.g., a new intervention), it must implicitly assume how that intervention would alter transmission, behavior, and surveillance assumptions that are often unstated, underidentified from the data, or simply absent from the model~\cite{dyaf015}. This motivates the need for true world models.

\section{World Models in Computational Epidemiology}

\subsection{Epidemics as Dynamical Systems with Intervention}
\vspace{-2mm}
We view epidemics as dynamical systems in which interventions are central drivers of system evolution rather than incidental covariates. In this framing, policies are actions applied over time, and humans are adaptive agents whose behavior responds to both disease conditions and policy choices~\cite{TALUKDER2024100292}. This creates a setting where decision-making must be counterfactual, i.e., reasoning about what would happen under alternative actions, not merely predictive. This perspective aligns with planning-as-inference views of decision-making, in which policies are evaluated by reasoning over future trajectories under alternative actions \cite{wood2022planning}. It also requires separating latent epidemic reality from the observed surveillance signals that imperfectly measure it, and it demands planning under uncertainty when both dynamics and observations can shift in response to interventions~\cite{Parag2025AsymmetricLimits}.







\subsection{Ingredients of an Epidemiological World Model}
\label{subsec:ingredients}
\vspace{-2mm}

We model an epidemic as a controlled, partially observed dynamical system in which (i) the epidemic state is latent, (ii) observations are imperfect and policy-dependent proxies of that state, and (iii) interventions affect transmission primarily through behavioral and social feedback. This formulation extends classical state-space models for infectious-disease surveillance by explicitly treating interventions as actions in a sequential decision process rather than as fixed covariates. We adopt the standard formulation of controlled, partially observed dynamical systems from planning and reinforcement learning, and instantiate it for epidemiological policy reasoning \cite{kaelbling1998planning}.

\paragraph{World model as a controlled state-space system.}
An epidemiological world model specifies a generative process over latent states, actions, and observations:
\begin{align}
a_t &\sim \pi_\psi(\cdot \mid h_t), 
\quad h_t = \phi(o_{1:t-1}, a_{1:t-1}), \label{eq:policy}\\
x_{t+1} &\sim P_\theta(x_{t+1}\mid x_t, a_t), \label{eq:transition}\\
o_t &\sim \Omega_\theta(o_t\mid x_t, a_{t-1}), \label{eq:observation}
\end{align}
where $x_t \in \mathcal{X}$ denotes the latent epidemic state, $o_t \in \mathcal{O}$ the observed surveillance signals, and $a_t \in \mathcal{A}$ the intervention applied at time $t$. The internal information state $h_t$ summarizes past observations and interventions and provides the information available for action selection under partial observability.


The encoder $\phi(\cdot)$ corresponds to an explicit belief state obtained via filtering, but need not be an exact Bayesian posterior. More generally, $\phi(\cdot)$ may be implemented as a learned recurrent or memory-based representation that implicitly encodes uncertainty, delayed effects, and long-range policy context, without assuming identifiability of the latent state.

\paragraph{Latent state.}
The epidemic state $x_t$ is entirely latent and need not be disentangled or semantically aligned with individual epidemiological variables. It can consist of structured components, continuous latent embeddings, or hybrid representations, provided it is sufficient to render the dynamics Markov under intervention. Conceptually, the state subsumes multiple interacting factors:
\begin{equation}
x_t \;=\; \big(x_t^{\text{epi}},\, x_t^{\text{imm}},\, x_t^{\text{net}},\, x_t^{\text{beh}},\, x_t^{\text{reg}}\big),
\end{equation}
where $x_t^{\text{epi}}$ captures true infections and susceptibility (including unobserved and asymptomatic cases), $x_t^{\text{imm}}$ denotes immunity and waning, $x_t^{\text{net}}$ captures the contact and mobility structure, $x_t^{\text{beh}}$ represents the behavioral and social responses (e.g., compliance, risk perception), and $x_t^{\text{reg}}$ encompasses slowly varying regime factors such as circulating variants, seasonality, or healthcare capacity. These quantities are unified as state variables because they are not directly observed yet jointly determine future epidemic evolution under intervention.

\paragraph{Dynamics.}
The transition model $P_\theta(x_{t+1}\mid x_t, a_t)$ governs how epidemic, behavioral, and regime components evolve in response to interventions. A defining feature of epidemiological dynamics is that actions affect transmission indirectly via behavior and contact patterns, inducing nonlinear and nonstationary effects:
\begin{equation}
P_\theta(x_{t+1}\mid x_t, a_t)
\;\approx\;
P_\theta\!\left(x_{t+1}\mid x_t,\, \eta(x_t^{\text{beh}}, a_t)\right),
\end{equation}
where $\eta(\cdot)$ denotes the induced response of population behavior and mixing. Regime components evolve on longer timescales, enabling the model to represent gradual drift and abrupt shifts, such as variant emergence, within a unified controlled dynamical system.

\paragraph{Observations.}
Observations $o_t$ correspond to quantities that are actually measured such as reported cases, hospitalizations, deaths, wastewater signals, or mobility traces, and are generated by a policy-dependent observation process:
\begin{equation}
o_t \sim \Omega_\theta(o_t \mid x_t, a_{t-1}).
\end{equation}
This process subsumes all mechanisms by which latent epidemic states become observable, including but not limited to testing practices, reporting pipelines, delays, measurement noise, and behavioral responses induced by interventions. Because policies affect both disease dynamics and surveillance itself, the relationship between $x_t$ and $o_t$ is endogenous to past actions. This separation between latent reality and observation is essential for reasoning under partial observability, delayed feedback, and systematic or strategic measurement bias.
\vspace{-1em}

\paragraph{Policies and actions.}
A policy $\pi_\psi$ maps the internal information state $h_t$ to interventions $a_t$. The action space includes non-pharmaceutical interventions, vaccination and testing strategies, risk communication, and enforcement intensity. Epidemiologically, actions are typically coarse and temporally extended, and their effects are mediated through social response rather than direct biological control.


\vspace{-2mm}
\section{Motivating Challenges}
\label{sec:challenges}
\vspace{-2mm}

We present three illustrative case studies that demonstrate why forecasting-centric approaches are insufficient for policy-relevant epidemiological reasoning. These studies are intentionally minimal and controlled. Rather than optimizing predictive accuracy, they are designed to expose structural failure modes that arise due to strategic behaviors, the nature of surveillance, and counterfactual intervention, settings that epidemiological world models are explicitly constructed to address. 
For each case study, we identify the problem, describe how a world-model perspective clarifies the failure mode, and enables more policy-relevant reasoning and representative results and findings.

\subsection{Epidemic Modeling under User Deception}
\label{subsec:misreporting}
\vspace{-2mm}


Epidemiological modeling increasingly relies on large-scale, crowdsourced behavioral data to track and forecast disease transmission dynamics in near real-time \cite{banerjee2022}. Beyond clinical surveillance 
records, self-reported information on vaccination uptake, mask usage, and adherence to social distancing has become central to assessing the impact of non-pharmaceutical interventions (NPIs) \cite{bff2025}.

\begin{wrapfigure}{r}{0.5\linewidth}
    \centering
    \vspace{-0.5em}
    \includegraphics[width=0.85\linewidth]{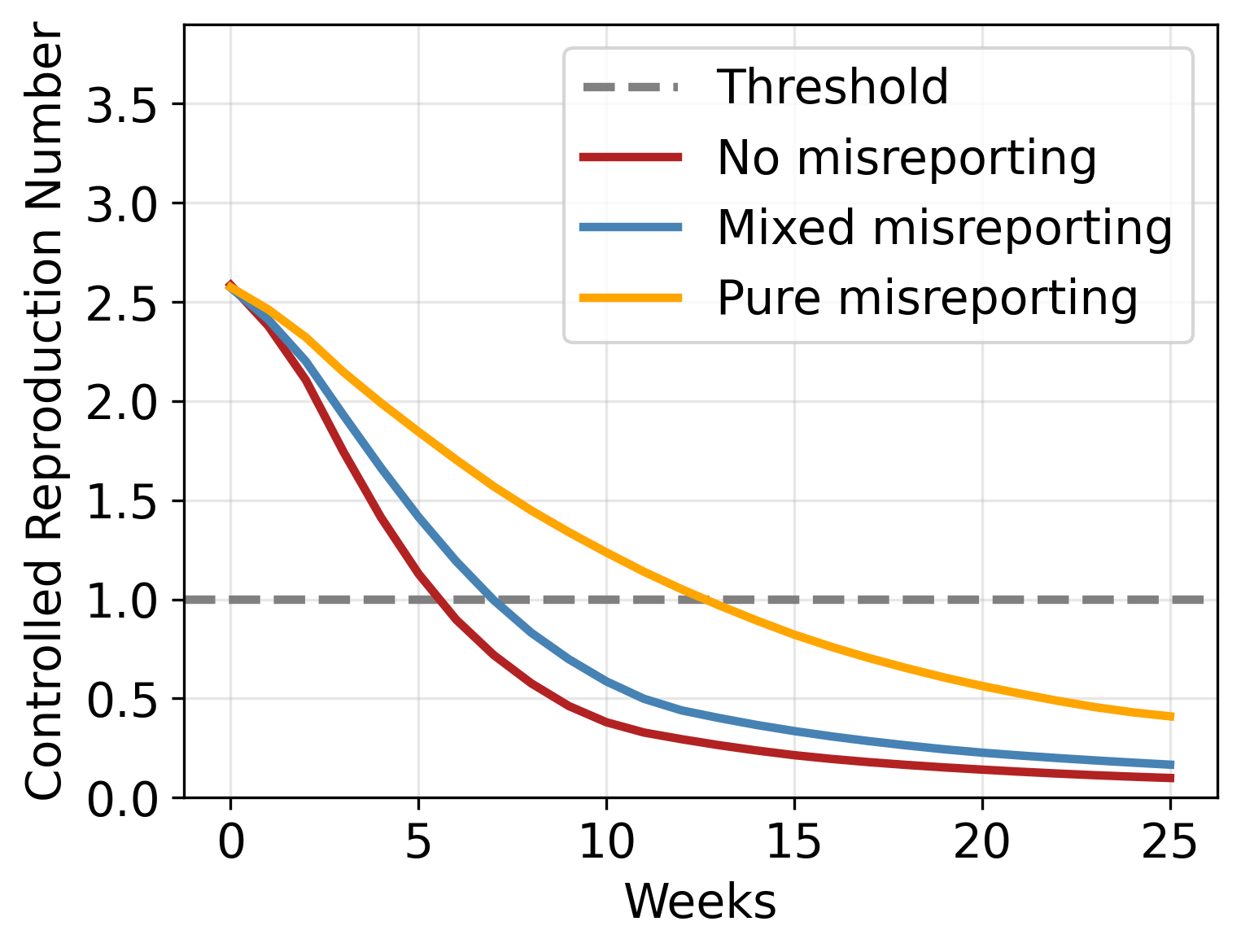}
    \vspace{-0.8em}
    \caption{Impact of strategic misreporting on epidemic control.}
    \label{fig:misreporting}
\end{wrapfigure}
This misalignment between reported and actual behavior can systematically bias parameter estimates and weaken NPIs \cite{WuNature2020,Pullano2021,MilanesiArxiv2023}.

Unfortunately, there are substantial and persistent gaps between self-reported and directly observed behaviors \cite{kenya2021,spatiotemporal}.
Even vaccination records, though more structured, exhibit persistent misreporting rates of 5–15\% due to reporting delays, incomplete registries, and misclassification~\cite{HURLEY2026128101,DALEY20242740,IRVING20096546}.
Beyond measurement error, self-reported epidemiological data are often strategically distorted. Individuals may misreport or withhold information due to testing avoidance, stigma, policy incentives, or distrust in public health authorities ~\cite{levy2022,shiman2023,su2026adversarial}. 

Figure~\ref{fig:misreporting} illustrates how strategic misreporting degrades epidemic control through biased observation of individual behavior under three regimes—no misreporting, mixed misreporting, and pure misreporting. Although all scenarios begin from the same initial transmission level, increasing degrees of misreporting systematically delay the reduction of the reproduction number below the threshold. This occurs because misreported vaccination and masking behavior leads the public health authority to overestimate compliance, resulting in weaker or delayed interventions despite persistent transmission. Consequently, misreporting acts as an endogenous distortion of the observation process that directly impairs control effectiveness rather than as random measurement noise.

However, most epidemiological models either ignore strategic misreporting~\cite{PatelJTB2005} or treat it as static parameter uncertainty (e.g.,~\cite{TannerMB2008}), and are evaluated primarily by predictive accuracy on reported cases. World models address misreporting by explicitly modeling how biased, incentive-driven observations are generated from latent epidemic states, allowing policies to be evaluated and optimized on inferred reality rather than on distorted reports~\cite{wm, hafner2020}.

\subsection{Epidemic Modeling under a Moving Target}
\label{subsec:timelag}



A large body of work has shown that epidemiological surveillance signals are inherently delayed proxies of underlying transmission dynamics. Delays arise from incubation periods, care-seeking behavior, testing, hospital admissions, and reporting pipelines, causing observed trends in confirmed cases, hospitalizations, and deaths to lag true infections by days to weeks \cite{Gostic2020,Pullano2021,Wu2020}. These delays have been shown to bias real-time inference, degrade forecast performance near turning points, and delay apparent responses to interventions in both influenza and COVID-19 surveillance \cite{Reich2019PNAS,Davies2020NHB}. 

\begin{wrapfigure}{l}{0.6\linewidth}
    \centering
\includegraphics[width=\linewidth]{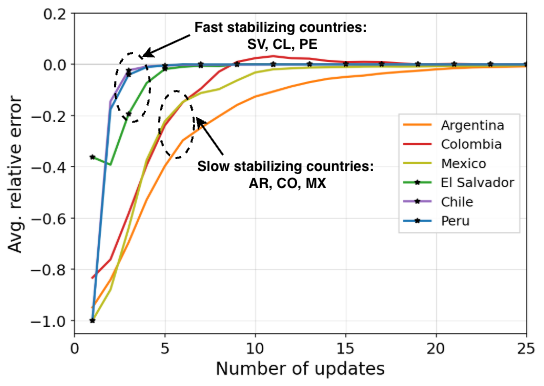}
    \caption{Dynamics of stabilization of reported ILI case counts across various countries.}
    \label{fig:lag}
\end{wrapfigure}
\vspace{2mm}

\vspace{2mm}
In Fig~\ref{fig:lag} we track reported ILI (influenza like illness) case counts in multiple countries of Latin America against the final, `resting', case counts after updates and corrections are incorporated over subsequent weeks. As can be seen, some countries are fast stabilizing (presumably, owing to the sophistication of their surveillance apparatus) and others are slower-stabilizing. 
The Centers for Disease Control and Prevention monitors influenza activity through a nationwide surveillance system in the United States that aggregates clinical, laboratory, and reporting data.
Despite its importance for decision-making, this system inherently lags real-time influenza transmission due to delays in symptom onset, care-seeking, and reporting~\cite{Mathis2024,pnas.ili}.
Incorporating these dynamics allows us to explicitly capture such observation dynamics into our world model.

Accurate infectious disease forecasting 
can support advance planning for vaccination clinics, targeted communication campaigns, and the distribution of antiviral medications, and may help guide mitigation measures such as school closures or cancellation of large gatherings during pandemics \cite{Reich2019PNAS, Mathis2024}. However, because forecasting models are trained to extrapolate delayed surveillance signals, their predictions remain conditioned on the same lagged and potentially distorted observations \cite{Gostic2020,Pullano2021}. As a result, forecasting alone cannot resolve the fundamental disconnect between observed data and the underlying epidemic state, motivating the need for world models that explicitly separate latent dynamics from the observation process and reason counterfactually about interventions.

\subsection{Counterfactual Analysis}
\label{subsec:counterfactual}

We next illustrate the necessity of world models for counterfactual policy analysis. In practice, the true number of infected individuals is rarely observable, and decision makers must rely on other (delayed) signals such as hospitalizations. Accordingly, we treat hospitalizations as the available ground-truth outcome and relate them to latent infections through a fixed infection--hospitalization ratio~\cite{Deng2025IHR}.

\begin{wrapfigure}{r}{0.55\linewidth}
    \centering
    \includegraphics[width=0.8\linewidth]{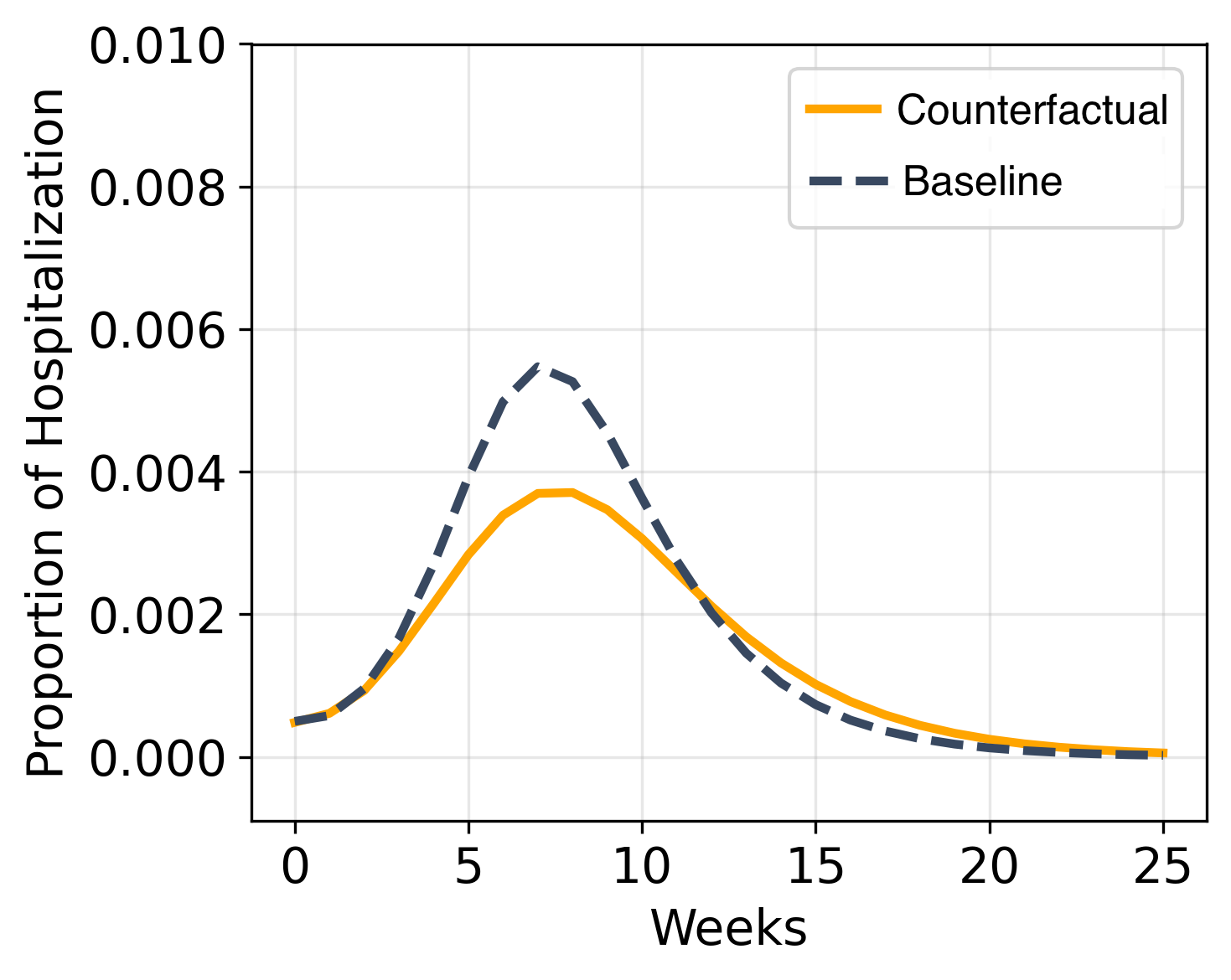}
    \vspace{-0.8em}
    \caption{Hospital admission trajectories under baseline and counterfactual interventions.}
    \label{fig:hosp}
\end{wrapfigure}

Figure~\ref{fig:hosp} shows the resulting hospitalization trajectories from a calibrated baseline and a counterfactual scenario with earlier vaccine eligibility and stricter masking mandates.
While the baseline scenario follows the realized course of the epidemic, the counterfactual intervention yields a lower and delayed hospitalization peak, corresponding to improved infection control. Crucially, this divergence arises despite identical observed histories prior to intervention, and cannot be recovered by forecasting approaches that condition only on the realized trajectory.

This example highlights a fundamental limitation of forecasting-centric methods: predicting future observations does not answer counterfactual questions about alternative actions. By contrast, world models enable intervention-aware simulation, allowing policy evaluation in terms of latent epidemic control rather than passive extrapolation of observed trends~\cite{alles2025latentactionworldmodels}.

\section{Framework}

\begin{figure}[htbp!] 
    \centering
    \vspace{-0.5em}
\includegraphics[width=0.99\textwidth]{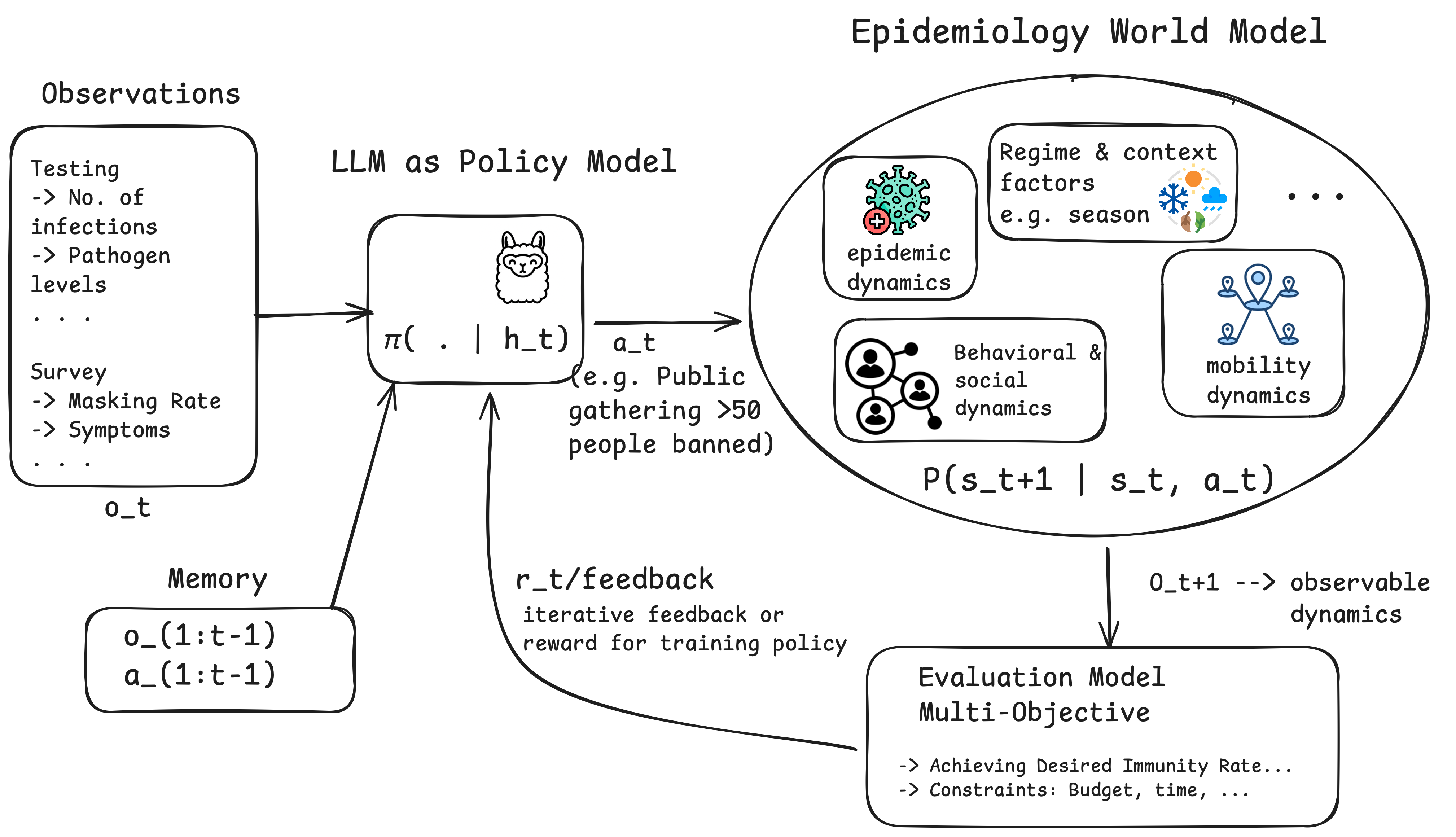}
    \caption{Epidemiological world model loop.}
    \vspace{-0.8em}
    \label{fig:schematic} 
\end{figure}





Having defined the components of an epidemiological world model, we now describe how these components are composed into a computational framework for policy reasoning. The framework is algorithm-agnostic: it does not prescribe how the world model is learned or how policies are optimized, but specifies the interaction loop through which candidate interventions are proposed, rolled out, evaluated, and refined. Figure~\ref{fig:schematic} illustrates this interaction loop, highlighting the separation between policy proposal, world-model simulation, observation generation, and feedback.

At each decision point, the framework maintains an internal information state $h_t$, as defined in Section~\ref{subsec:ingredients}. Given $h_t$, candidate policies propose interventions, which are evaluated by rolling out the epidemiological world model forward in time. The resulting latent trajectories and synthetic observations are scored using policy-relevant objectives. This loop can be executed once for scenario analysis or iterated to refine policies. 

\subsection{Policy models}
\vspace{-2mm}
Policies can be rule-based, learned, or hybrid. In particular, large language models (LLMs) can act as \emph{policy models} that map the internal information state $h_t$ into candidate interventions. In this role, LLMs provide a strong prior over plausible decision rules and intervention structures, but do not determine outcomes directly: all candidate actions are evaluated through world-model rollouts.

\subsection{World model rollouts}
\label{subsec:rollout}
\vspace{-2mm}

Given a candidate intervention sequence $a_{t:t+H}$, the world model simulates forward trajectories of the latent epidemic state:
\begin{equation}
    x_{t+1:t+H} \sim P_\theta(\cdot \mid x_t, a_{t:t+H}),
\end{equation}
where $H$ denotes the planning horizon. The use of a horizon reflects the fact that epidemiological interventions are temporally extended and exhibit delayed effects: changes in transmission, hospitalizations, and mortality typically manifest over weeks rather than instantaneously. Evaluating policies therefore requires reasoning over multi-step futures, including transient dynamics, delayed surveillance signals, and cumulative outcomes such as healthcare overload.

Rollouts propagate epidemiological, behavioral, and regime components, allowing interventions to affect transmission through contact structure and compliance. This enables the framework to capture both short-term responses (e.g., behavioral adaptation following an intervention) and longer-term consequences (e.g., immunity accumulation or variant-driven regime shifts).

In parallel, the observation model generates synthetic surveillance signals:
\begin{equation}
    o_{t+1:t+H} \sim \Omega_\theta(\cdot \mid x_{t+1:t+H}, a_{t:t+H}),
\end{equation}

making explicit how different policies shape both the underlying epidemic and the data that would be observed. This distinction is critical in epidemiology, where observed trends may reflect surveillance effects rather than true changes in transmission.

\subsection{Evaluation and Iterative Refinement}
\label{subsec:evaluation}
\vspace{-2mm}

Each rollout is evaluated using outcome measures aligned with epidemiological decision-making. These may include health outcomes (e.g., cumulative infections, deaths, hospital overload), social and economic costs, and equity considerations. Evaluation is typically multi-objective and can involve constraints rather than a single scalar score (e.g., avoiding ICU capacity violations). Importantly, evaluation operates on latent trajectories as well as synthetic observations, allowing policies to be assessed on their true epidemic impact rather than solely on predicted reported cases.

Evaluation outputs provide feedback to the policy model. In the simplest case, candidate interventions are compared and filtered based on dominance or constraint satisfaction. More generally, evaluation can induce scalar or vector-valued reward signals derived from multi-objective criteria, enabling reinforcement learning, bandit-style updates, or preference-based optimization. When policies are represented by language-based models, structured evaluation feedback can be incorporated through iterative prompting or refinement.

The framework is agnostic to the specific update mechanism, but makes explicit how intervention outcomes feed back into subsequent policy proposals. This iterative loop—policy proposal $\rightarrow$ world model rollout $\rightarrow$ evaluation $\rightarrow$ refinement—constitutes the computational core by which epidemiological world models support counterfactual reasoning and planning under uncertainty, beyond forecasting-centric approaches.

\subsection{Learning the World Model and Policy}
\label{sec:learning}
\vspace{-2mm}

We learn the world model from trajectories of observations and interventions,
\((o_{1:T}, a_{1:T})\). Under partial observability, we maintain a belief state
\begin{equation}
h_t = f_{\phi}(h_{t-1}, o_t, a_{t-1}),
\end{equation}
together with controlled latent dynamics and an observation model:
\begin{align}
x_{t+1} &\sim P_{\theta}(x_{t+1}\mid x_t, a_t), \\
o_t &\sim \Omega_{\theta}(o_t\mid x_t, a_{t-1}),
\end{align}

\vspace{-.8mm}
where $\phi$ denotes the parameters of the belief-state encoder and $\theta$ denotes the jointly learned parameters of the transition and observation models. 
We train the world model by maximizing a sequential variational objective:
\begin{equation}
\mathcal{L}_{\text{WM}}
=
\sum_{t=1}^{T}
\mathbb{E}_{q_{\phi}(x_t)}
\big[
\log \Omega_{\theta}(o_t \mid x_t)
\big]
-
\beta\,\mathrm{KL}\!\left(
q_{\phi}(x_{1:T})
\;\|\;
p_{\theta}(x_{1:T}\mid a_{1:T})
\right),
\end{equation}
which encourages latent states to explain observations while remaining consistent with controlled dynamics.
After training, policies are optimized using model-based rollouts:
\begin{equation}
\pi^{*}
=
\arg\max_{\pi}
\mathbb{E}
\left[
\sum_{\tau=t}^{t+H} r(x_{\tau}, a_{\tau})
\right],
\end{equation}
where trajectories are simulated using the learned world model. This enables counterfactual evaluation of interventions beyond forecasting.


\section{Experimental Results}
\vspace{-2mm}
\subsection{Settings}
\vspace{-2mm}
We instantiate the framework using COVID-19 surveillance data from all 50 US states (January 2021--late 2022). State features include weekly epidemiological signals (hospitalizations per 100k, confirmed cases, vaccination rates), static demographic and healthcare-capacity attributes, and 13-dimensional policy actions from the Oxford COVID-19 Government Response Tracker (OxCGRT)~\cite{hale2021global} encoded on a 0--4 ordinal scale. The world model is CovidLLM, a LLaMA-2-7B~\cite{touvron2023llama} base fine-tuned via SFT with LoRA~\cite{hu2022lora}; epidemiological context is serialized into a structured prompt alongside the proposed policy action, with a GRU encoder providing temporal context over multi-week sequences. The model predicts next-week hospitalization trend as an ordinal classification over five classes. We evaluate two policy model families, GPT-4o-mini~\cite{achiam2023gpt} and Qwen-7B\cite{bai2023qwen}, in closed-loop rollouts across all 50 states starting from September 5, 2022, with a planning horizon of $H{=}6$ weeks. For Qwen-7B, we additionally evaluate two optimization strategies: iterative feedback and GRPO. Policy alignment is measured as the percentage of proposed actions matching historically realized OxCGRT interventions; hospitalization reduction is the mean percentage decrease in hospitalizations per 100k over the horizon.

\subsection{Results}
\vspace{-2mm}

\paragraph{Policy evaluation.}
GPT-4o-mini achieves 44.8\% alignment with historically realized interventions across all 50 states and 13 policy dimensions, correctly capturing the directionality and coarse structure of non-pharmaceutical interventions. Qwen-7B obtains 30.3\% alignment, a notable gap that suggests smaller open-weight models are less reliably calibrated to the scale conventions of the OxCGRT action space, likely due to weaker instruction-following and reduced exposure to policy-structured data during pretraining or reasoning capabilities. GPT-4o-mini drives a mean hospitalization reduction of 46.5\% over the 6-week horizon, while Qwen-7B achieves 39.1\%, both exceeding the 37.1\% reduction observed under historically realized interventions (9.09 $\to$ 5.67 per 100k).

\begin{table}
\small
\centering
\caption{Policy evaluation and optimization results across all 50 US states ($H{=}6$ weeks). Policy alignment is the percentage of proposed actions matching historically realized interventions. Hosp.\ reduction is the mean percentage decrease in hospitalizations per 100k over the rollout horizon.}
\label{tab:results}
\vspace{2mm}
\begin{tabular}{llcc}
\toprule
\textbf{Category} & \textbf{Method} & \textbf{Policy Alignment\,(\%)\,$\uparrow$} & \textbf{Hosp.\ Reduction\,(\%)\,$\uparrow$} \\
\midrule
\multirow{2}{*}{Policy Eval.} & GPT-4o-mini & 44.8 & $-$46.5 \\
 & Qwen-7B & 30.3 & $-$39.1 \\
\midrule
\multirow{2}{*}{Policy Optim.} & Iterative Feedback & 49.5 & $-$48.7 \\
 & GRPO & 42.2 & $-$46.1 \\
\bottomrule
\end{tabular}
\end{table}
\vspace{-2mm}

\paragraph{Policy optimization (Qwen-7B).}
The policy optimization conditions use world-model rollouts as a signal to refine the Qwen-7B policy model's system prompt beyond zero-shot prompting. The \emph{iterative feedback} method~\cite{agrawal2025gepa} observes the full sequence of hospitalization outcomes across all rollout steps and uses these observations to rewrite the policy model's system prompt, steering it toward lower-burden intervention sequences in subsequent iterations. This approach achieves 49.5\% alignment and a hospitalization reduction of 48.7\% (the highest across all conditions), indicating that prompt-level refinement guided by simulated epidemic outcomes can elicit more epidemiologically coherent intervention recommendations. The \emph{GRPO} treats the negative end-of-horizon hospitalization count as a scalar reward and optimizes the policy model's action distribution directly via group relative policy gradient updates over on-policy rollouts in the world model, achieving 42.2\% alignment and a 46.1\% hospitalization reduction.

\vspace{-2mm}
\section{Discussion}
\vspace{-2mm}
While our framework provides a principled formulation of epidemiological world models, several challenges remain. First, latent epidemic states are inherently difficult to identify from noisy, delayed, and policy-dependent observations, leading to ambiguity between true infections, behavior, and reporting processes. This is further compounded by limited and heterogeneous data availability across regions, making multi-modal integration (e.g., cases, hospitalizations, mobility, surveys) both necessary and nontrivial. In addition, world models introduce significant computational overhead due to latent-state inference, multi-step rollouts, and policy optimization, particularly in multi-region or agent-based settings. Designing realistic and tractable policy representations also remains an open problem, as interventions are often high-level, constrained, and shaped by social and institutional factors that are difficult to encode.

Looking forward, several directions can strengthen this line of work. Incorporating mechanistic priors and causal structure may improve identifiability and interpretability of latent states, while advances in multi-modal learning and missing-data handling can make models more robust to real-world surveillance conditions. Scalable training and planning methods, including hierarchical and approximate approaches, will be essential for deployment at population scale. Additionally, developing richer evaluation protocols beyond forecasting accuracy, such as counterfactual validity and policy effectiveness, will be critical for assessing real-world impact. Ultimately, integrating world models with multi-agent, multi-region settings and adaptive behavioral dynamics offers a promising path toward more reliable and actionable epidemic decision support.

\vspace{-2mm}
\section{Conclusion}
\vspace{-2mm}
In closing, we advocate a shift from forecasting-centric epidemic modeling to an explicit world-model view that treats epidemics as controlled, partially observed dynamical systems with endogenous surveillance and adaptive human behavior.
While our studies are minimal, they surface concrete failure modes that forecasting alone cannot resolve and illustrate the kinds of robustness-oriented analyses that epidemiological world models make possible. We believe that epidemiological world models can serve as reliable substrates for intervention design and stress-testing.

\vspace{-2mm}
\section*{Acknowledgements}
\vspace{-2mm}
This work is supported in part by US National Science Foundation grants 
CCF-1918770,  
IIS-2509636, 
IIS-2312794, 
DBI-2412389, 
and CNS-2225511. 
Any opinions, findings, and conclusions or recommendations expressed in this material are those of the author(s) and do not necessarily reflect the views of the sponsor(s).

\bibliography{iclr2026_conference}
\bibliographystyle{iclr2026_conference}


\end{document}

%% file: math_commands.tex
\usepackage{amsmath,amsfonts,bm}

\def\eqref#1{equation~\ref{#1}}

\def\1{\bm{1}}

\DeclareMathAlphabet{\mathsfit}{\encodingdefault}{\sfdefault}{m}{sl}
\SetMathAlphabet{\mathsfit}{bold}{\encodingdefault}{\sfdefault}{bx}{n}

